\documentclass{article}
\usepackage[preprint]{neurips_2026}

\usepackage[utf8]{inputenc}
\usepackage[T1]{fontenc}
\usepackage{hyperref}
\usepackage{url}
\usepackage{booktabs}
\usepackage{amsfonts}
\usepackage{nicefrac}
\usepackage{microtype}
\usepackage{xcolor}
\usepackage{pifont}
\usepackage{amsmath}
\usepackage{enumitem}
\usepackage{algorithm}
\usepackage{algpseudocode}
\usepackage{graphicx}
\usepackage{subcaption}
\usepackage{multirow}
\usepackage{colortbl}
\hypersetup{hidelinks}

\newcommand{\rev}[1]{#1}
\definecolor{wcTeal}{HTML}{16856B}
\definecolor{wcAmber}{HTML}{C06B16}
\definecolor{wcMuted}{HTML}{8A8F98}
\definecolor{wcTableHeader}{HTML}{E8E8E8}
\definecolor{wcTableHighlight}{HTML}{DDF4F8}
\newcommand{\yes}{\textcolor{wcTeal}{\ding{51}}}
\newcommand{\na}{\textcolor{wcMuted}{-}}
\newcommand{\up}{\textcolor{wcTeal}{\ensuremath{\uparrow}}}
\newcommand{\down}{\textcolor{wcAmber}{\ensuremath{\downarrow}}}

\title{WorldCraft: Persistent Object-State Control in Camera-Navigable Video World Models}

\author{
  Bohai Gu\thanks{Work done as an intern at Tencent AI Technology Center, Tencent Video.}$^{1,2}$,\quad
  Taiyi Wu$^{2}$,\quad
  Yueyang Yuan$^{3}$,\quad
  Jian Liu$^{1}$,\quad
  Xiaocheng Lu$^{1}$,\quad
  Dazhao Du$^{1}$,\\
  Jie Zhang$^{1}$,\quad
  Jinxiang Lai$^{1}$,\quad
  Shuai Yang$^{4}$,\quad
  Xiaotong Zhao$^{2}$,\quad
  Alan Zhao$^{2}$,\quad
  Song Guo\thanks{Corresponding author.}$^{1}$ \\[6pt]
  \normalfont\small
  $^{1}$The Hong Kong University of Science and Technology \quad
  $^{2}$AI Technology Center, Tencent Video, Tencent \\
  $^{3}$Wuhan University \quad
  $^{4}$Peking University
}

\begin{document}

\maketitle

\begin{abstract}
Recent video world models enable interactive camera navigation but provide limited control over individual objects. We study \emph{persistent object-state control} in autoregressive video world models, where an object-trajectory action must remain effective across viewpoint changes, generation chunks, and intervals of temporary invisibility. This setting introduces an action-memory conflict because autoregressive memory records the last observed object state, whereas an intervening action may update that state off-screen. We present \textbf{WorldCraft}, \rev{the first camera-navigable autoregressive video world model to jointly compose camera navigation and object-trajectory control.} Given a user-selected object and a 2D motion path, Normalized World Trajectory (NWT) anchors the path in a normalized world coordinate system and projects it into each camera view, producing a consistent object-action signal throughout the autoregressive rollout. \rev{Trajectory-Anchored State Persistence (TASP) combines persistent NWT guidance with selective memory refresh to maintain action-updated object states through off-screen intervals and re-entry.} A pathway-selective LoRA introduces object control while retaining the pretrained camera controller. \rev{WorldCraft achieves accurate composable camera-object control and preserves camera fidelity. We further find that it maintains action-updated object states across extended intervals of complete target invisibility without off-screen trajectory supervision.}
\end{abstract}

\section{Introduction}

A world model~\citep{dreamer,worldmodels} is a learned simulator that predicts future states given the current state and an action.
Recent video-based world models realize this idea directly in pixel space, such as Genie~3~\citep{genie3} and WorldPlay~\citep{worldplay} have made impressive progress on \emph{camera-level} interaction: users can navigate viewpoints and the model generates coherent visual continuations.
Yet their action space stops at the camera.
Real interaction is inherently object-centric.
A robot must predict the consequence of pushing a cup along a tabletop~\citep{daydreamer, unisim}. A driving simulator must model a pedestrian entering the road~\citep{drivedreamer, gaia}. An interactive game must allow users to move entities rather than only the viewpoint.
In all these cases, the action is a continuous trajectory attached to a specific object.
Without such object-level actions, world models are more like passive scene observers than manipulable environments.

Object-level actions in an autoregressive world model differ from trajectory guidance within a fixed generation window~\citep{wanmove,draganything,motionctrl}. Fixed-window methods represent a trajectory as a sequence of visible spatial conditions. \rev{We define \textbf{persistent object-state control} as the ability to maintain an action-induced state transition across viewpoint changes and autoregressive chunk boundaries, including intervals in which the target is outside the current observation.} This requirement creates two coupled problems. First, \emph{observation-dependent coordinates}: screen-space motion entangles object displacement with camera ego-motion, so the same intended object action changes under a moving viewpoint. \rev{Second, \emph{off-screen object-state updates} create an action-memory conflict. While the target is fully outside the current view, an object-level trajectory may continue moving it from its last observed location to a new state. Autoregressive visual memory, however, remains anchored to the pre-exit state. Upon re-entry, the model must render the same object at its trajectory-specified location rather than restore the stale location stored in memory.} 

\rev{We introduce \textbf{WorldCraft}, a framework for persistent object-state control in camera-navigable autoregressive video world models.} Given a user-selected object and a 2D motion path, WorldCraft composes object-trajectory control with continuous camera navigation. Normalized World Trajectory (NWT) anchors the user path in a normalized reference coordinate system and re-projects it under each camera pose, separating intended object motion from camera-induced image displacement across viewpoints and autoregressive chunks. \rev{After each chunk, NWT updates its visual anchor and depth estimate from the latest generated observation to limit accumulated re-projection drift.} \rev{Trajectory-Anchored State Persistence (TASP) uses NWT as global object-state guidance during off-screen intervals and filters stale pre-exit memories at re-entry, preserving object appearance and scene context while enforcing the updated position. We implement trajectory adaptation with Spatial-Pathway LoRA (SP-LoRA), which equips the backbone with object-level trajectory control while preserving its pretrained camera-navigation capability.}

Table~\ref{tab:comparison} contrasts WorldCraft with fixed-window motion-control methods and camera-navigable world models. \rev{To our knowledge, WorldCraft is the first camera-navigable autoregressive video world model to jointly compose camera navigation and object-trajectory control within a single rollout. Importantly, its training data contain no trajectories that move a controlled object while it is fully off-screen. We therefore evaluate \emph{zero-shot} transfer to this unseen regime using a dedicated benchmark, where the trajectory updates an invisible target and the returning camera tests whether it reappears at the specified location. Together, NWT propagates the trajectory-updated spatial state and TASP reconciles it with visual memory at re-entry, maintaining the update through up to four invisible generation chunks.} Our contributions are:

\begin{table}[t]
  \caption{\textbf{Comparison of control and generation settings.} ``Off-screen object-state update'' denotes changing a controlled object's state through an object-level action while the target is fully outside the camera view.}
  \label{tab:comparison}
  \centering
  \small
  \setlength{\tabcolsep}{3.2pt}
  \resizebox{\linewidth}{!}{
  \begin{tabular}{llcccc}
    \toprule
    Method & Setting & Camera & Object & AR rollout & \shortstack{Off-screen object-\\state update} \\
    \midrule
    \multicolumn{6}{l}{\textit{Motion-controlled generation and editing}} \\
    DragAnything~\citep{draganything} (ECCV) & Fixed-window generation & \na & \yes & \na & \na \\
    Wan-Move~\citep{wanmove} (NeurIPS) & Fixed-window generation & \na & \yes & \na & \na \\
    MotionCtrl~\citep{motionctrl} (SIGGRAPH) & Fixed-window generation & \yes & \yes & \na & \na \\
    Free-Form Motion Control~\citep{shuai2025free} (ICCV) & Fixed-window generation & \yes & \yes & \na & \na \\
    Edit-by-Track~\citep{lee2026generative} (CVPR) & Fixed-window editing & \yes & \yes & \na & \na \\
    \midrule
    \multicolumn{6}{l}{\textit{Interactive video world models}} \\
    Yume~\citep{mao2025yume}(Arxiv) & Camera-navigable world model & \yes & \na & \yes & \na \\
    Matrix-Game~2.0~\citep{matrixgame2}(Arxiv)  & Camera-navigable world model & \yes & \na & \yes & \na \\
    GameCraft~\citep{gamecraft}(Arxiv)  & Camera-navigable world model & \yes & \na & \yes & \na \\
    Genie~3~\citep{genie3} & Camera-navigable world model & \yes & \na & \yes & \na \\
    WorldPlay~\citep{worldplay} (ICML 2026) & Camera-navigable world model & \yes & \na & \yes & \na \\
    \midrule
    \textbf{WorldCraft (ours)} & Persistent object-state control & \yes & \yes & \yes & \yes \\
    \bottomrule
  \end{tabular}
  }
\end{table}

\begin{enumerate}[leftmargin=*]

\item \textbf{Joint camera-object control in autoregressive world models.}
\rev{To our knowledge, WorldCraft is the first camera-navigable autoregressive video world model to jointly compose camera navigation and object-trajectory control within a single rollout.}

\item \textbf{Autoregressive camera-decoupled trajectory control.}
\rev{NWT decouples object motion from camera-induced displacement and iteratively updates its anchor and depth across autoregressive chunks.}

\item \textbf{Zero-shot persistent object-state control.}
\rev{Without training trajectories that move fully off-screen targets, NWT and TASP jointly propagate trajectory-updated states through invisible intervals and recover controlled objects at the specified locations upon re-entry.}

\end{enumerate}

\section{Related work}

\paragraph{Interactive video world models and state-evolution benchmarks.}
Recent video world models make controllable simulation in pixel space increasingly practical. GameNGen~\citep{GameNGen} and GameGen-X~\citep{gamegenx} demonstrate action-conditioned game video generation~\citep{yu2025gamefactory}, while Yume~\citep{mao2025yume}, GameCraft~\citep{gamecraft}, Matrix-Game~2.0~\citep{matrixgame2}, and WorldPlay~\citep{worldplay} advance long-horizon camera navigation and scene revisitation. WorldPlay is our direct backbone because it combines camera-action conditioning with autoregressive visual memory. Such memory can preserve previously observed scene content during camera excursions. However, these models do not expose an object-level action that changes a selected object's state while it is outside the current view. WorldCraft adds this object-level state-update capability while retaining camera navigation and scene memory. 
\rev{MIND~\citep{mind} evaluates memory consistency and action control through closed-loop scene revisitation, while STEVO-Bench~\citep{stevobench} tests whether naturally evolving processes continue under camera lookaway. These benchmarks establish unobserved state evolution as a distinct challenge for video world models. Our OSS benchmark targets a complementary trajectory-conditioned setting: a user trajectory updates a selected object's position while it is fully outside the view, and the same autoregressive camera rollout must recover that object at the action-specified state upon re-entry.}

\paragraph{Camera and object motion control.}
Trajectory-guided methods provide object control through motion paths~\citep{dragnuwa,draganything} or displaced first-frame features~\citep{wanmove}. MotionCtrl~\citep{motionctrl} jointly controls camera poses and object trajectories in fixed-window generation, Free-Form Motion Control~\citep{shuai2025free} uses explicit 6D camera and object poses, and Edit-by-Track~\citep{lee2026generative} edits a source video with paired 3D point tracks. These methods establish camera-object control for fixed-duration generation or editing under inputs ranging from 2D paths to explicit 3D constraints.
\rev{WorldCraft instead studies persistent object-state control in a camera-navigable autoregressive world model. From a user-selected object and 2D motion path, NWT produces view-dependent conditions that evolve with the camera and generated state across chunks. The action must compose with camera navigation, remain effective through long-horizon memory, and update a fully invisible target. Our quantitative comparisons accordingly use trajectory methods with a matched 2D-path interface and camera-controlled world models that support autoregressive long-video evaluation.}

\section{Method}

\begin{figure}[t]
  \centering
  \includegraphics[width=1\linewidth]{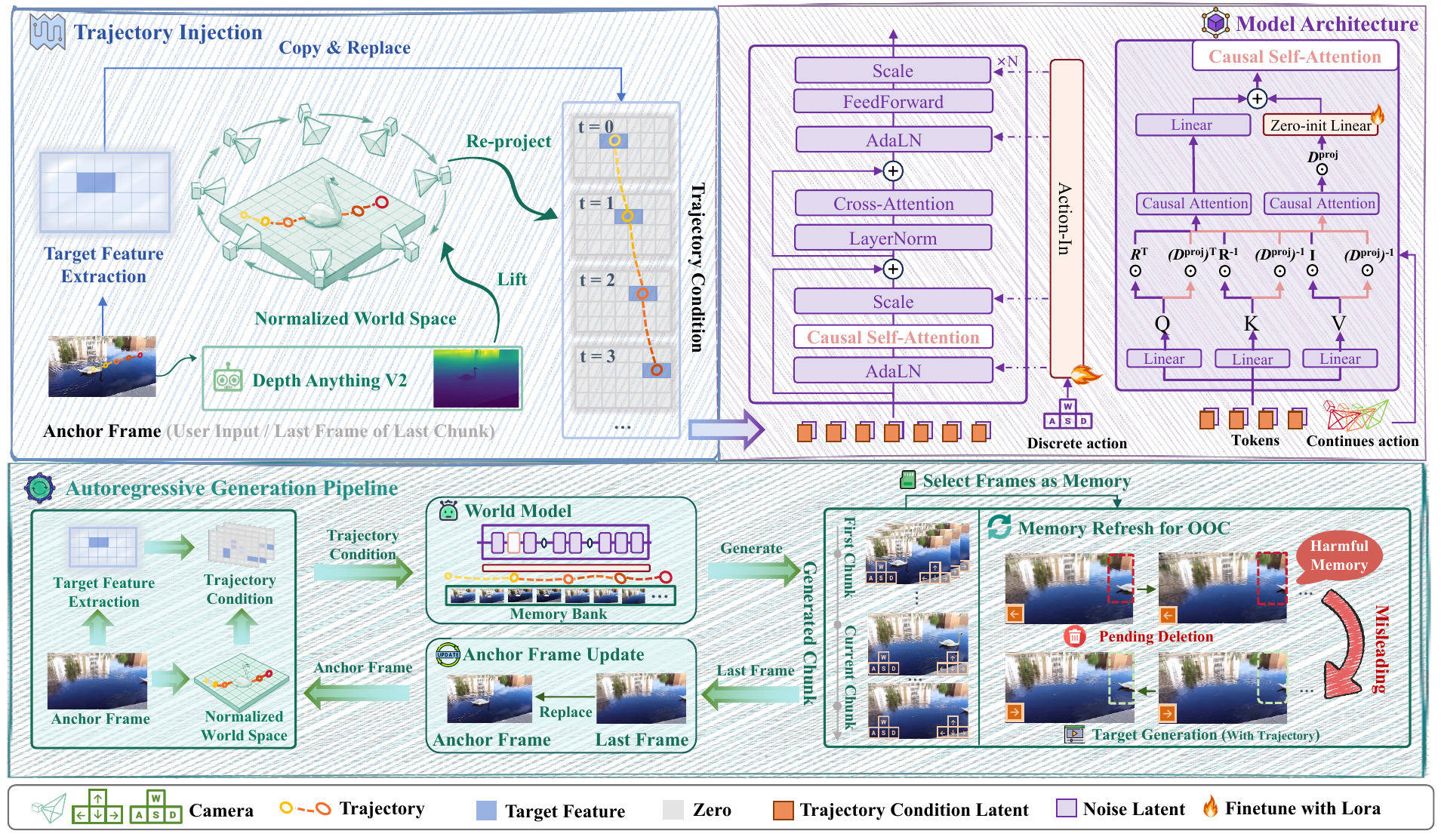}

\caption{\textbf{WorldCraft overview.}
\textbf{(Top-left)} WorldCraft lifts a user-specified 2D trajectory into a camera-decoupled normalized world space and re-projects it into per-frame trajectory conditions under the given camera actions.
\textbf{(Top-right)} The trajectory and camera controls are injected through a lightweight pathway-selective LoRA on the spatial-control pathway, while the backbone attention and MLP layers remain frozen.
\textbf{(bottom)} During autoregressive generation, WorldCraft updates the anchor frame and memory bank across chunks, and refreshes outdated memories to support long-horizon out-of-camera object reasoning.
}
  \label{fig:overview}
  \vspace{-3mm}
\end{figure}

WorldCraft converts a transient trajectory condition into persistent object-state control for an autoregressive video world model. Figure~\ref{fig:overview} summarizes the pipeline. After describing the backbone and in-context trajectory injection (\S\ref{sec:prelim}-\ref{sec:injection}), \rev{we present NWT for camera-decoupled trajectory representation and iterative updates across chunks (\S\ref{sec:nwt}), followed by TASP for reconciling off-screen object-state updates with autoregressive visual memory (\S\ref{sec:tasp}). We then describe Spatial-Pathway LoRA for backbone-preserving trajectory adaptation (\S\ref{sec:splora}).} \rev{The method proceeds from action representation to state persistence: NWT determines the trajectory coordinates across views and chunks, and TASP reconciles the resulting action state with autoregressive memory. Spatial-Pathway LoRA provides a lightweight implementation for adapting the backbone to this trajectory condition.}

\subsection{Preliminaries and notation}
\label{sec:prelim}

\rev{WorldCraft builds on WorldPlay~\citep{worldplay}, an autoregressive video DiT~\citep{dit,vit} based on HunyuanVideo-1.5~\citep{hunyuanvideo15}. Its input concatenates a 32-channel noisy latent $\mathbf{z}_t$, a 32-channel image condition $\mathbf{c}_\text{img}$, and a task mask $\mathbf{m}$. WorldPlay generates video chunk by chunk, conditions camera actions through the action encoder $\texttt{action\_in}$ and per-block ProPE projections $\texttt{prope\_proj}^{(l)}$~\citep{li2025camerasrelativepositionalencoding}, and attends to cached key-value memories from previous chunks~\citep{causvid}. We denote frame-$t$ camera intrinsics and world-to-camera extrinsics by $\mathbf{K}_t \in \mathbb{R}^{3\times3}$ and $\mathbf{E}_t \in \mathrm{SE}(3)$, and perspective projection by $\pi(\mathbf{K},\mathbf{E},\mathbf{P})$.}

\subsection{In-context trajectory conditioning}
\label{sec:injection}

We inject trajectory information by replacing the image-conditioning channels (32-64) with first-frame latent features displaced to the target positions.
Given the first-frame latent $\mathbf{z}_0 \in \mathbb{R}^{C \times 1 \times H \times W}$ and $N$ point trajectories $\{\mathbf{p}_t^{(n)}\}_{n,t}$, the trajectory condition $\hat{\mathbf{c}}_\text{traj}$ is:
\begin{equation}
\hat{\mathbf{c}}_\text{traj}[\,:,\, t,\, h_t^{(n)},\, w_t^{(n)}\,] \;\leftarrow\; \mathbf{z}_0[\,:,\, 0,\, h_0^{(n)},\, w_0^{(n)}\,], \quad \forall\; n,\, t
\label{eq:replace}
\end{equation}
where $(h_t^{(n)}, w_t^{(n)})$ is the latent-space coordinate of track~$n$ at frame~$t$. Unassigned positions remain zero.
The model input becomes $[\,\mathbf{z}_t \;;\; \hat{\mathbf{c}}_\text{traj} \;;\; \mathbf{m}\,]$, preserving full compatibility with the pretrained PatchEmbed~\citep{hunyuanvideo15}.
This design yields an informative prior even \emph{before} trajectory-specific training: displaced first-frame features serve as positional cues that the base model's spatial understanding partially decodes, producing coarse trajectory following at zero shot.

\subsection{Normalized World Trajectory}
\label{sec:nwt}
At inference, the user clicks the target object in the initial frame and sketches a 2D path specifying its desired motion. NWT acts as a lightweight \emph{action compiler}, converting this input into a camera-aware control signal that remains meaningful beyond the current image. A raw screen-space path entangles the object's intended displacement with apparent motion caused by camera ego-motion. We therefore anchor the trajectory to the first-frame camera coordinate system and dynamically re-project it under the current camera pose. \rev{For autoregressive generation, NWT further updates the visual anchor and target depth after each chunk, limiting accumulated re-projection drift while retaining the same user-specified world-space trajectory.}

\paragraph{Definition.}
Given a user-specified screen trajectory $\{\mathbf{p}_t^\text{user}\}$ and a reference depth $d$, we first lift each point to a normalized coordinate on the first-frame reference plane:
\begin{equation}
\mathbf{q}_t =
\begin{pmatrix}
(x_t - c_x) / f_x \\
(y_t - c_y) / f_y
\end{pmatrix},
\label{eq:world}
\end{equation}
where $f_x, f_y, c_x, c_y$ are the first-frame intrinsic parameters.
Intuitively, $\mathbf{q}_t$ describes the object's position on a first-frame-anchored reference plane, rather than its instantaneous screen-space location.
At each generation step, we re-project this anchored coordinate into the current camera view:
\begin{equation}
\mathbf{p}_t^\text{anchored}
=
\pi\Bigl(
\mathbf{K}_t,\;
\mathbf{E}_t \mathbf{E}_0^{-1}
\cdot
\mathrm{lift}(\mathbf{q}_t, d)
\Bigr),
\label{eq:reproject}
\end{equation}
where $\mathbf{E}_t$ denotes the world-to-camera extrinsic matrix and $\mathrm{lift}(\mathbf{q}_t,d)$ maps the normalized coordinate to a 3D point on the reference depth plane.
This re-projection automatically folds camera-induced pixel displacement into the trajectory signal consumed by the model.

\paragraph{Composable camera-object control.}
Trajectories extracted from videos by point tracking are screen-space observations that naturally entangle object motion with camera motion:
\begin{equation}
\mathbf{p}_t = \pi\bigl(\mathbf{K}_t,\; \mathbf{E}_t,\; \mathbf{P}_\text{world}(t)\bigr),
\label{eq:entangle}
\end{equation}
At inference time, however, users typically specify the desired object motion without compensating for the camera trajectory, leading to a mismatch if the raw user path is used directly as screen-space conditioning.
NWT closes this gap by anchoring the user trajectory in the first-frame coordinate system and re-projecting it under the current camera pose.
As a result, user-specified object motion and model-driven camera motion can be composed without requiring the user to manually anticipate camera-induced screen displacement.

\paragraph{Depth estimation and iterative anchor refinement.}
Eq.~\ref{eq:reproject} requires the object depth $d$, which we initialize by querying the monocular depth map~\citep{depthanythingv2} at the user's initial click position:
$d_0 = \mathcal{D}(\mathbf{I}_0)[\mathbf{p}^\text{user}_0]$.
In long autoregressive rollouts, however, keeping both the depth and the visual anchor fixed to the first frame can accumulate projection errors across chunks.
We therefore apply \emph{iterative anchor refinement}: after each generated chunk, we update the autoregressive anchor to the latest reliable frame, re-estimate the object depth from that frame, and use the updated anchor-depth pair for subsequent re-projection.
This forms a closed-loop correction that keeps the trajectory condition aligned with the generated video, reducing geometric drift while preserving the normalized world-space trajectory as the global control signal.

\begin{algorithm}[t]
\caption{Inference with Normalized World Trajectory}
\label{alg:world-space}
\begin{algorithmic}[1]
\Require First frame $\mathbf{I}_0$; user screen trajectory $\{\mathbf{p}^\text{user}_t\}_{t=0}^{T}$; camera pose sequence $\{\mathbf{E}_t\}_{t=0}^{T}$; first-frame intrinsics $\mathbf{K}_0$; chunk size $C$.
\Ensure Generated video $\{\mathbf{I}_t\}_{t=1}^{T}$.
\Statex \textit{\# Stage 1: lift user trajectory into world space (once).}
\State $d \gets \mathcal{D}(\mathbf{I}_0)[\mathbf{p}^\text{user}_0]$ \Comment{depth at initial click}
\For{$t = 0, \ldots, T$}
    \State $\mathbf{q}_t \gets \bigl((x^\text{user}_t - c_x)/f_x,\; (y^\text{user}_t - c_y)/f_y\bigr)$
\EndFor
\Statex \textit{\# Stage 2: autoregressive generation with iterative depth refinement.}
\For{chunk $k = 1, \ldots, \lceil T/C \rceil$}
    \For{$t$ in chunk $k$}
        \State $\mathbf{p}^\text{anchored}_t \gets \pi\bigl(\mathbf{K}_t,\; \mathbf{E}_t \cdot \mathbf{E}_0^{-1} \cdot \mathrm{lift}(\mathbf{q}_t, d)\bigr)$
    \EndFor
    \State $\mathbf{I}_{\text{chunk } k} \gets \mathrm{WorldModel}\bigl(\mathbf{I}_{<\text{chunk } k},\, \{\mathbf{p}^\text{anchored}_t\},\, \{\mathbf{E}_t\}\bigr)$
    \State $d \gets \mathrm{RefineDepth}(\mathbf{I}_{\text{chunk } k},\, \mathbf{p}^\text{anchored}_{t_\text{last}},\, d)$
\EndFor
\State \Return $\{\mathbf{I}_t\}_{t=1}^{T}$
\end{algorithmic}
\end{algorithm}

\subsection{Trajectory-Anchored State Persistence}
\label{sec:tasp}

A world model is partially observed: leaving the camera view removes an object from the observation, not from the simulated world. \rev{An \emph{off-screen object-state update} occurs when an object-level trajectory moves a fully invisible target away from its last observed location. Autoregressive memory still encodes the pre-exit position, creating a conflict with the trajectory-specified state required at re-entry.} \rev{TASP reconciles these two sources of state: NWT carries the updated object position through the invisible interval, while selective filtering prevents stale pre-exit memory from restoring the old position. The remaining memory continues to provide appearance, identity, and scene context.}

\textit{(i) Persistent trajectory state.}
\rev{The normalized trajectory $\{\mathbf{q}_t\}$ remains defined independently of object visibility. During the camera-away interval $(t_0,t_1)$, the trajectory continues updating the target state across autoregressive chunks even though the target is not observed. When the camera returns at $t_1$, Eq.~\ref{eq:reproject} maps the current trajectory state into the returning view, providing the action-updated re-entry position.}

\textit{(ii) Pre-exit memory filtering.}
When the camera turns away from an object at time $t_0$ and returns at $t_1$, the autoregressive memory $\mathcal{M}_{t_0}$ holds a \emph{frozen state snapshot}: its keys and values encode the object at its pre-departure screen location.
At $t_1$, attention retrieval confidently reproduces that stale location even if the trajectory has moved the object elsewhere.

We resolve this with \emph{pending deletion with dynamic mask}: at each re-entry chunk, we identify memory frames in the pre-exit zone (the last $k$ temporal latent before off-screen happens) and mask them from the retrieval set if their FOV similarity with the current chunk exceeds a threshold $\tau$:
\begin{equation}
\mathrm{sim}_\text{FOV}(\mathbf{V}_f,\, \mathbf{V}_\text{cur}) > \tau \;\Longrightarrow\; f \notin \mathcal{M}
\label{eq:preexit}
\end{equation}
where $\mathbf{V}_f$ and $\mathbf{V}_\text{cur}$ are the view matrices of memory frame $f$ and the current chunk's first frame.
Frames outside the pre-exit zone or with dissimilar FOV are retained, preserving the appearance prior.
\rev{The two mechanisms are complementary. The persistent trajectory supplies the updated location at re-entry, while pre-exit filtering suppresses stale positional evidence and retains the remaining memory as an appearance and scene prior.}

\subsection{Backbone-preserving trajectory adaptation}
\label{sec:splora}

\rev{Camera actions and object trajectories differ in scope but both regulate where visual content appears. A camera action produces a scene-level transformation, whereas a trajectory moves a selected instance. We therefore hypothesize that trajectory adaptation should reuse the backbone's spatial-control pathway rather than broadly modify attention and feature-mixing layers. Full-parameter trajectory fine-tuning in Appendix~\S\ref{sec:splora-ablation} supports this hypothesis: relative weight changes concentrate on the action encoder and ProPE projections, which change by $10$-$25{\times}$ more than attention and MLP layers.} \rev{Motivated by this observation, \emph{Spatial-Pathway LoRA}~\citep{lora} applies low-rank updates only to \texttt{action\_in} and the per-block \texttt{prope\_proj} layers, while freezing the remaining backbone. The resulting ${\sim}50$M-parameter adapter ($0.6\%$ of the backbone) adds trajectory control while limiting interference with the pretrained camera controller. Appendix~\S\ref{sec:splora-ablation} gives the complete weight-change analysis, update formulation, layer-selection ablation, and training-strategy comparison.}

\section{Experiments}

\subsection{Implementation details}
\label{sec:setup}

\noindent\textbf{Base model and training.} We build on WorldPlay~\citep{worldplay}, an 8B-parameter video world model based on HunyuanVideo-1.5~\citep{hunyuanvideo15}. WorldCraft follows a three-stage progressive schedule. Stage~0 performs real-video domain adaptation. Stage~1 learns trajectory control from static-camera data using BI attention and SP-LoRA. Stage~2 transfers this capability to dynamic-camera data using AR attention. 
Details are provided in Appendix~\S\ref{sec:training}.

\noindent\textbf{Evaluation data.}
All evaluation splits are disjoint from training. The \textbf{Trajectory Accuracy (TA)} set contains 50 static-camera clips with SAM~2 object masks~\citep{sam2} and CoTracker trajectories~\citep{cotracker}. The \textbf{Camera Fidelity (CF)} set contains 50 dynamic-camera clips without object actions, using per-latent intrinsics and extrinsics estimated by ViPE~\citep{vipe}. The \textbf{Composable} set contains 45 clips evenly divided among small ($<15^\circ$), medium ($15$-$45^\circ$), and large ($\geq45^\circ$) camera rotations, with both camera and object controls. The \textbf{Off-Screen State (OSS)} set contains 50 held-out 253-frame rollouts with mirrored camera-away-and-return paths. The trajectory updates the fully invisible target before the reversed camera path reveals its trajectory-specified re-entry state; we evaluate one and four consecutive invisible chunks. No training trajectory moves a controlled target while fully off-screen.

\noindent\textbf{Evaluation horizon and metrics.}
We evaluate trajectory control with Trajectory Error (TE), the mean CoTracker~\citep{cotracker} L2 pixel error between the tracked object and specified path, and visual quality with VBench++~\citep{vbenchpp,vbench}, PSNR/SSIM~\citep{ssim}, and LPIPS~\citep{lpips}. Camera control is measured by rotational, translational, and combined Relative Pose Error (RPE) after Sim(3) Umeyama alignment~\citep{umeyama1991}. For OSS, \textbf{RTE\textsubscript{f}} measures final re-entry error relative to the trajectory-specified location, and \textbf{RID\textsubscript{pre}} measures DINO~\citep{dinov2} identity similarity to the pre-exit target. Background-only \textbf{MEt3R}~\citep{asim2026met3rmeasuringmultiviewconsistency} and relative-depth consistency (\textbf{DepthC}) are computed on matched outbound/return views with the target's swept region excluded.

\begin{table}[t]
  \caption{\textbf{Trajectory control} under static camera (61 frames, 50 clips). All methods share the same first frame and trajectory condition. Best in bold.}
  \centering
  \setlength{\tabcolsep}{5pt}
  \renewcommand{\arraystretch}{1.12}
  \resizebox{\linewidth}{!}{
  \begin{tabular}{@{}l cccc ccc c@{}}
    \toprule
    & \multicolumn{4}{c}{\textbf{Visual Quality}}
    & \multicolumn{3}{c}{\textbf{VBench++}} & \\
    \cmidrule(lr){2-5}\cmidrule(lr){6-8}
    \rowcolor{wcTableHeader}
    \textbf{Method} & \textbf{PSNR}\up & \textbf{SSIM}\up & \textbf{LPIPS}\down & \textbf{DINO}\up
                    & \textbf{SubjC}\up & \textbf{BgC}\up & \textbf{Temp}\up & \textbf{TE}\down \\
    \midrule
    DragAnything~\citep{draganything} (ECCV 2024) & 15.97 & 0.600 & 0.468 & 0.777 & 0.896 & 0.913 & 0.938 & 39.86 \\
    Wan-Move~\citep{wanmove} (NeurIPS 2025) & 16.42 & 0.592 & 0.375 & 0.782 & 0.927 & 0.943 & 0.985 & 44.08 \\
    \rowcolor{wcTableHighlight}
    \textbf{WorldCraft (ours)} & \textbf{17.23} & \textbf{0.616} & \textbf{0.363} & \textbf{0.807} & \textbf{0.942} & \textbf{0.945} & \textbf{0.989} & \textbf{38.90} \\
    \bottomrule
  \end{tabular}
  }
  \label{tab:main}
\end{table}

\subsection{Quantitative results}
\label{sec:main-results}

\paragraph{Evaluation protocols.}
\rev{We report numerical comparisons only when the conditioning interface and evaluation protocol can be matched. For trajectory control, DragAnything~\citep{draganything}, Wan-Move~\citep{wanmove}, and WorldCraft receive the same initial frame and 2D target trajectory (Table~\ref{tab:main}). For camera fidelity, we compare against the camera-controlled world models Yume~\citep{mao2025yume}, Matrix-Game~2.0~\citep{matrixgame2}, GameCraft~\citep{gamecraft}, and WorldPlay~\citep{worldplay} under the same camera-action sequences (Table~\ref{tab:camera}). 

Additional composable-control and DROID results are provided in Appendix~\S\ref{sec:supp-joint}.}

\noindent\textbf{Trajectory control quality.}
Under static camera with an identical first frame and object-trajectory condition, WorldCraft achieves the lowest trajectory error (TA) while simultaneously producing the best pixel fidelity (PSNR/SSIM~\citep{ssim}/LPIPS~\citep{lpips}), semantic consistency (DINO), and VBench++ consistency scores across all 50 TA clips (Table~\ref{tab:main}). 

\noindent\textbf{Camera fidelity.}
On camera-only inputs, WorldCraft retains the camera-control capability of the base model. At 61 frames, its RPE\textsubscript{rot} is $0.131$, compared with $0.120$ for WorldPlay, and far below the next-best external baseline ($0.252$). At 253 frames, a horizon $4{\times}$ longer than the main protocol, WorldCraft further reduces the error to $0.123$, outperforming WorldPlay ($0.130$). These results show that object-level trajectory control preserves camera fidelity and slightly improves long-horizon stability (Table~\ref{tab:camera}).

\begin{table}[t]
  \caption{\textbf{Camera fidelity} on camera-only input. WorldCraft preserves camera accuracy at 61 frames and outperforms all methods at the 253-frame extended horizon.}
  \label{tab:camera}
  \centering
  \small
  \setlength{\tabcolsep}{3.5pt}
  \renewcommand{\arraystretch}{1.12}
  \resizebox{\linewidth}{!}{
  \begin{tabular}{@{}l ccc ccc !{\hspace{2pt}\vrule width 0.7pt\hspace{2pt}} ccc@{}}
    \toprule
    & \multicolumn{6}{c}{\textbf{Short-term (61 frames)}}
    & \multicolumn{3}{c}{\textbf{Long-term (253 frames)}} \\
    \cmidrule(lr){2-7}\cmidrule(lr){8-10}
    \rowcolor{wcTableHeader}
    \textbf{Method} & \textbf{RPE\textsubscript{rot}}\down & \textbf{RPE\textsubscript{trans}}\down & \textbf{RPE\textsubscript{cam}}\down
                    & \textbf{PSNR}\up & \textbf{SSIM}\up & \textbf{LPIPS}\down
                    & \textbf{RPE\textsubscript{rot}}\down & \textbf{RPE\textsubscript{trans}}\down & \textbf{RPE\textsubscript{cam}}\down \\
    \midrule
    Yume~\citep{mao2025yume} & 0.261 & 0.0143 & 0.0169 & 12.39 & 0.2931 & 0.5718 & 0.374 & 0.0247 & 0.0285 \\
    Matrix-Game~2.0~\citep{matrixgame2} & 0.342 & 0.0137 & 0.0196 & 12.96 & 0.3235 & 0.5326 & 0.162 & 0.0243 & 0.0261 \\
    GameCraft~\citep{gamecraft} & 0.252 & \textbf{0.0130} & \textbf{0.0157} & 12.42 & 0.2861 & 0.5529 & 0.198 & 0.0243 & 0.0265 \\
    WorldPlay (base)~\citep{worldplay} & \textbf{0.120} & 0.0155 & 0.0165 & 13.77 & 0.3434 & 0.4700 & 0.130 & 0.0262 & 0.0276 \\
    \rowcolor{wcTableHighlight}
    \textbf{WorldCraft (ours)} & 0.131 & 0.0161 & 0.0170 & \textbf{13.95} & \textbf{0.3474} & \textbf{0.4621} & \textbf{0.123} & \textbf{0.0225} & \textbf{0.0233} \\
    \bottomrule
  \end{tabular}
  }
  \vspace{-4pt}
\end{table}

\begin{figure*}[t]
\centering
\includegraphics[width=1\linewidth]{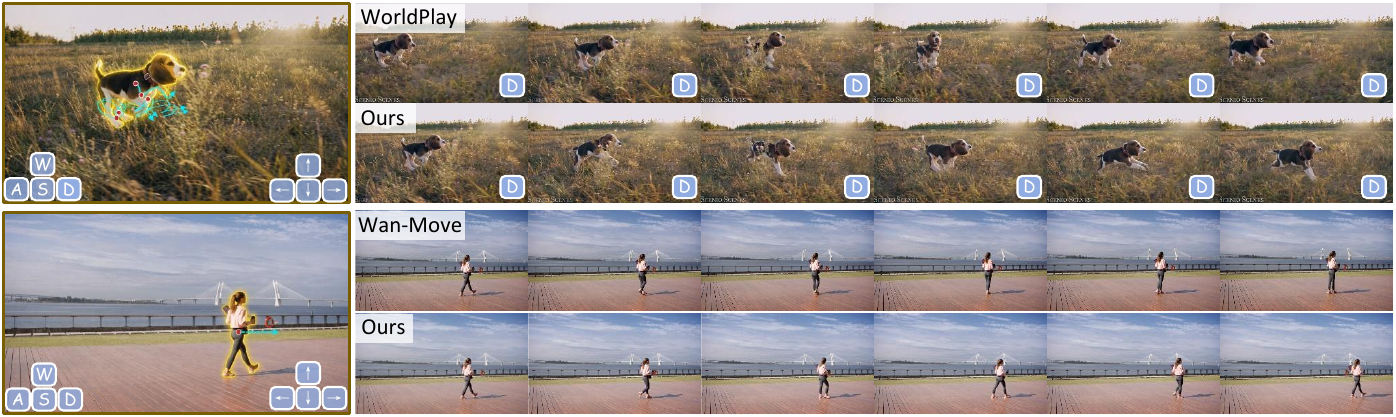}

\caption{\textbf{Qualitative comparison of trajectory control.} \textbf{WorldCraft} achieves precise and composable controllability, jointly controlling camera motion and target-object trajectories.}
\label{fig:qual_4way}
\end{figure*}

\begin{figure*}[t]
\centering
\includegraphics[width=1\linewidth]{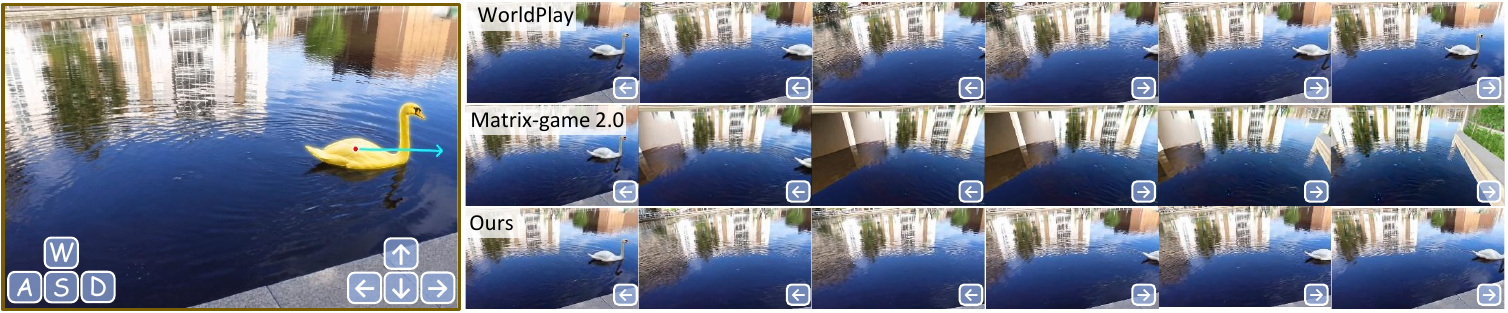}

\caption{\textbf{Long-horizon comparisons with off-camera motion.}
Given the same initial frame and camera actions, the goose moves right while the camera pans left and then returns. \textbf{WorldCraft} maintains scene consistency and, via TASP, recovers the goose at the correct off-camera-updated position when it re-enters view, whereas baselines either lose scene consistency or cannot track the off-camera object state.}
\vspace{-3mm}

\label{fig:qual_ooc}
\end{figure*}

\begin{table}[t]
\vspace{-1mm}
  \caption{\textbf{Trajectory-conditioned off-screen object-state propagation} on the 50-video OSS benchmark (253 frames). (a) Full TASP under a fourfold increase in target invisibility. (b) Matched memory ablation at four off-screen chunks.}
  \label{tab:offscreen}
  \centering
  \scriptsize
  \setlength{\tabcolsep}{3.2pt}
  \resizebox{\linewidth}{!}{
  \begin{tabular}{lcccccccc}
    \toprule
    Setting & TE$_\text{mean}$\down & RTE$_\text{f}$\down & RID$_\text{pre}$\up & MEt3R\down & DepthC\down & RPE$_\text{rot}$\down & RPE$_\text{trans}$\down & RPE$_\text{cam}$\down \\
    \midrule
    \multicolumn{9}{l}{\textit{(a) Robustness to off-screen duration (Full TASP)}} \\
    1 off-screen chunk & 38.48 & 32.70 & 0.805 & 0.173 & 0.0665 & 0.126 & 0.0165 & 0.0194 \\
    4 off-screen chunks & 40.03 & 35.11 & 0.717 & 0.196 & 0.0755 & 0.137 & 0.0182 & 0.0217 \\
    \midrule
    \multicolumn{9}{l}{\textit{(b) Action-memory reconciliation at 4 off-screen chunks}} \\
    Default WorldPlay memory & 52.10 & 48.20 & 0.618 & 0.202 & 0.0765 & 0.139 & \textbf{0.0171} & \textbf{0.0215} \\
    \textbf{TASP selective reconciliation} & \textbf{40.03} & \textbf{35.11} & \textbf{0.717} & \textbf{0.196} & \textbf{0.0755} & \textbf{0.137} & 0.0182 & 0.0217 \\
    \bottomrule
  \end{tabular}
  }
\end{table}

\subsection{Qualitative results}
\label{sec:qualitative}

\begin{figure}[b]
\centering
\includegraphics[width=1\linewidth]{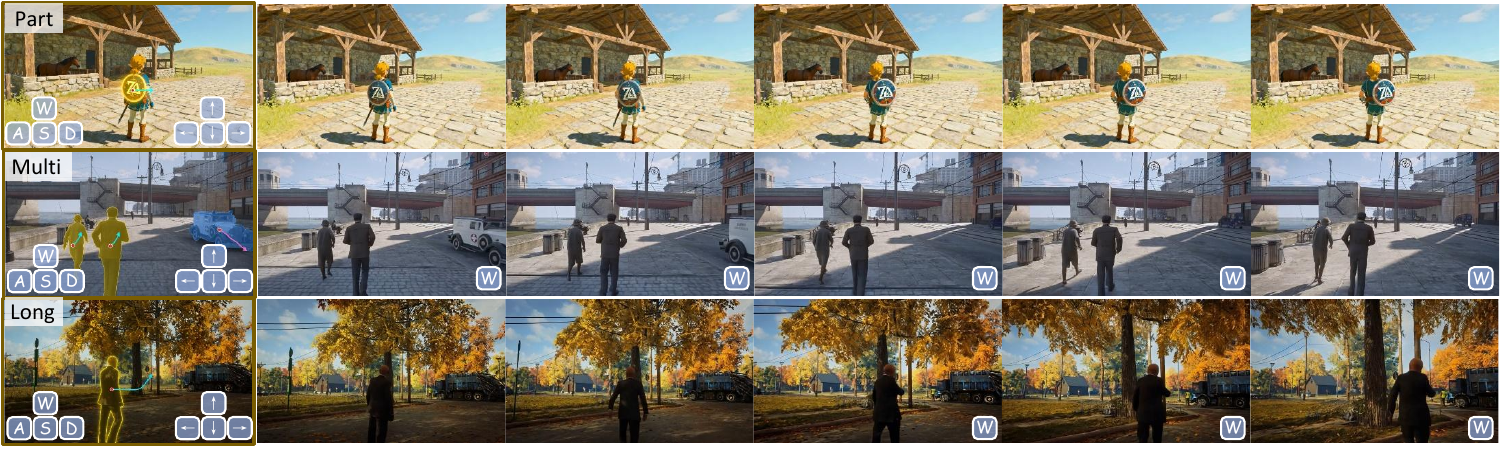}

\caption{\textbf{Extended capabilities.} \textbf{Part}: part-level control-the shield follows the trajectory while the body stays still. \textbf{Multi}: multi-object control-three objects steered simultaneously along independent trajectories. \textbf{Long}: $253$-frame autoregressive rollout with long trajectory (${\sim}10.5$\,s at 24\,fps).}
\label{fig:part_long_demo}
\vspace{-2mm}
\end{figure}

We present qualitative comparisons along three axes: Trajectory control (Figure~\ref{fig:qual_4way}), long-horizon camera rollout with off-camera demonstration (Figure~\ref{fig:qual_ooc}), and part-level control, multi-object control, long-term control ($253$-frame) of WorldCraft in Figure~\ref{fig:part_long_demo}.

\noindent\textbf{Trajectory control.}
Figure~\ref{fig:qual_4way} compares WorldCraft with representative baselines. WorldPlay produces plausible scenes but does not support precise object-level trajectories. Under sparse trajectory signals, WorldCraft follows the prescribed path throughout the rollout while preserving the target and scene.

\begin{table}[t]
 \vspace{-1mm}
  \caption{\textbf{NWT representation ablation.} Trajectory error on the composable set (61 frames, 45 clips) across camera-rotation magnitudes. World-space trajectories with iterative depth refinement perform best, especially under large rotations.}
  \label{tab:repr-ablation}
  \centering
  \small
  \setlength{\tabcolsep}{8pt}
  \begin{tabular}{l ccc}
    \toprule
    Representation & TE (small rot.)~\down & TE (mid rot.)~\down & TE (large rot.)~\down \\
    \midrule
    Pixel space (raw user traj.)                       & 35.79          & 40.61          & 45.33          \\
    World space + single-shot depth                           & 33.82          & 37.69          & 41.28          \\
    \textbf{World space + iterative depth }      & \textbf{30.63} & \textbf{32.10} & \textbf{34.65} \\
    \bottomrule
  \end{tabular}
   \vspace{-3mm}
\end{table}

\noindent\textbf{Off-screen state propagation.}
Figure~\ref{fig:qual_ooc} visualizes the action-memory conflict quantified in Table~\ref{tab:offscreen}. The goose moves right while the camera pans left, remains fully outside the observation, and later returns along the mirrored camera path. Camera-only baselines can revisit the scene but cannot execute the object action. WorldCraft instead carries the trajectory-conditioned state through the invisible interval and renders the same object at its updated position upon re-entry. The model receives no matched supervision for the off-screen/re-entry regime, while the intended motion remains specified by the input trajectory.

WorldCraft also supports (i)~\emph{Part-level control} (Figure~\ref{fig:part_long_demo}, top) and  (ii)~\emph{Multi-object control } simultaneously (Figure~\ref{fig:part_long_demo}, middle), even in (iii)~\emph{Long-horizon generation} (Figure~\ref{fig:part_long_demo}, bottom), extending to $253$ frames (${\sim}10.5$\,s) with composable camera-object control.

\subsection{Zero-shot off-screen object-state propagation}
\label{sec:offscreen-results}

The OSS benchmark tests whether object control learned from visible trajectories transfers to fully unobserved motion. Because no training trajectory moves a controlled target while it is fully off-screen, this evaluation is zero-shot with respect to the tested state-update regime. With NWT and TASP, increasing target invisibility from one to four chunks changes full-rollout TE from $38.48$ to $40.03$ and final re-entry error from $32.70$ to $35.11$ (Table~\ref{tab:offscreen}(a)). Identity and background consistency are more sensitive, so our robustness claim is limited to the evaluated four-chunk range.

Table~\ref{tab:offscreen}(b) isolates the contribution of TASP under four invisible chunks. Both variants receive the same NWT trajectory state and differ only in memory handling. Default WorldPlay memory retains stale pre-exit evidence, whereas TASP reconciles memory with the trajectory-updated state. TASP reduces final re-entry error from $48.20$ to $35.11$ ($27.2\%$), improves re-entry identity from $0.618$ to $0.717$, and largely preserves background geometry and camera control. Together, the duration study and matched ablation show that NWT and TASP generalize visible-object control to zero-shot off-screen state propagation.

\subsection{Ablation studies}
\label{sec:ablation}

\noindent\textbf{Normalized world-space trajectory.}
\rev{Under the same click-and-sketch interface, we isolate two design choices within NWT:
(i) pixel-space versus world-space coordinates, and
(ii) single-shot versus iterative monocular depth estimation.}
Table~\ref{tab:repr-ablation} reports trajectory error across three camera-rotation magnitudes.
Compared with raw pixel-space conditioning, world-space trajectories consistently improve trajectory accuracy, showing that anchoring trajectories in 3D space better composes object motion with camera.
Iterative depth refinement provides an additional gain, especially under large camera rotations, where repeated re-estimation helps correct projection drift accumulated during autoregressive rollout.

\noindent\textbf{Backbone-preserving adaptation.}
Full fine-tuning slightly reduces TE from $38.90$ to $37.20$ but increases rotational RPE from $0.131$ to $0.237$, whereas conventional Q/K/V$+$MLP LoRA uses ${\sim}4\times$ more trainable parameters and performs worse on both metrics. Spatial-Pathway LoRA therefore provides the best control-preservation trade-off; complete layer and curriculum ablations are reported in Appendix~\S\ref{sec:splora-ablation}.

\section{Conclusion}

We introduced WorldCraft for persistent object-state control in camera-navigable autoregressive video world models. \rev{The central problem is maintaining an action-induced state transition as observations change and visual memory retains stale evidence.} NWT represents a lightweight 2D trajectory as a camera-aware action across viewpoints and chunks. TASP preserves useful visual context while suppressing memories that conflict with the updated object position, and SP-LoRA retains the inherited camera controller. \rev{Without matched off-screen training supervision, WorldCraft propagates object state across up to four invisible chunks and recovers the target at its updated location.} These results extend interactive video world models from viewpoint navigation to persistent object-state control under partial observation.

{\small
\bibliographystyle{plainnat}

\bibliography{references}
}

\clearpage

\appendix

\section{Training and checkpoint details}
\label{sec:training}

Trajectory control and camera navigation rely on the same spatial-control pathway identified in \S\ref{sec:splora}.
Na\"ively training trajectory control therefore risks \emph{catastrophic interference} with the base model's camera capabilities.

\paragraph{Checkpoint initialization and three-stage pipeline.}
\rev{WorldPlay provides distinct bidirectional (BI) and autoregressive (AR) checkpoints. Stage~0 initializes from the 8B HunyuanVideo-1.5-based WorldPlay-BI checkpoint for real-domain adaptation, and Stage~1 retains the resulting BI backbone for static-camera trajectory learning. Stage~2 loads the corresponding WorldPlay-AR base weights for dynamic-camera autoregressive training and transfers the Spatial-Pathway LoRA learned in Stage~1. The base checkpoint therefore matches the attention regime at each stage, while the trajectory adapter remains continuous from Stage~1 to Stage~2.} We set the LoRA rank to 32 and otherwise follow the training configuration of WorldPlay. All experiments are conducted on 8 NVIDIA H200 GPUs.

\begin{table}[t]
  \centering
  \small
  \caption{Training schedule and checkpoint initialization. Batch size is reported per GPU, with gradient accumulation over eight steps.}
  \label{tab:training-schedule}
  \resizebox{\linewidth}{!}{
  \begin{tabular}{cllcccc}
    \toprule
    Stage & Attention/data & Base initialization & Updated parameters & Steps & LR & Frames \\
    \midrule
    0 & BI / real-domain & WorldPlay-BI & Full backbone & 2K & $5\!\times\!10^{-7}$ & 97 \\
    1 & BI / static-camera & Stage-0 BI backbone & Spatial-Pathway LoRA only & 40K & $5\!\times\!10^{-6}$ & 97 \\
    2 & AR / dynamic-camera & WorldPlay-AR & Transferred Spatial-Pathway LoRA only & 60K & $5\!\times\!10^{-6}$ & 97 \\
    \bottomrule
  \end{tabular}
  }
\end{table}

\noindent
\textbf{Stage~0} adapts WorldPlay-BI to the real-video domain through full-parameter fine-tuning at a low learning rate. No trajectory conditioning is used ($\texttt{trajectory\_rate}{=}0$), so this stage does not introduce camera-trajectory interference.
\textbf{Stage~1} freezes the Stage-0-adapted BI backbone and trains Spatial-Pathway LoRA on static-camera trajectory data. Static cameras remove the screen-space entanglement of Eq.~\ref{eq:entangle}, yielding a clean trajectory$\,\to\,$object-motion mapping.
\textbf{Stage~2} loads the WorldPlay-AR base checkpoint, keeps its backbone frozen, attaches the Spatial-Pathway LoRA learned in Stage~1, and continues optimizing that adapter on dynamic-camera data with AR attention. Stages~1 and~2 use a per-GPU batch size of one and gradient accumulation over eight steps. The AR-native initialization isolates the adapter transfer from changes in the backbone attention regime.

\section{Scalable data curation pipeline}
\label{sec:data}

\begin{figure}[t]
  \centering
  \includegraphics[width=\linewidth]{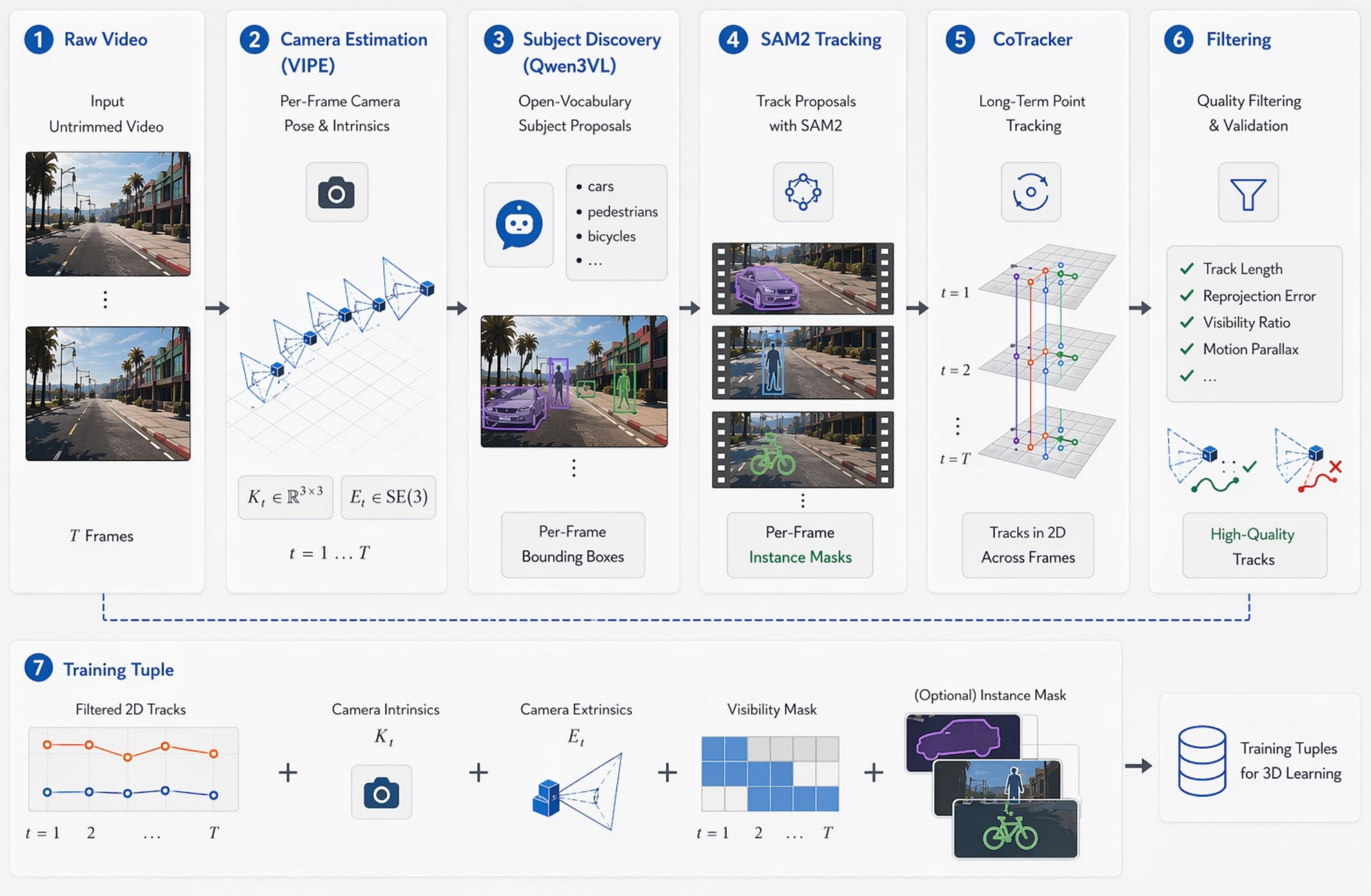}
  \caption{Automatic data curation pipeline. Given unlabeled video, we extract camera parameters, discover the salient moving subject, track it with SAM2 to obtain per-frame masks, and run CoTracker to produce multi-point trajectories. Physical-plausibility filters remove degenerate samples. The pipeline adapts to two data sources with complementary strengths: WISA-80K contributes diverse real-world scenes via a fully automatic VLM-guided discovery; SpatialVID-HQ provides metric camera annotations that we combine with a novel tracklet-based subject selection.}
  \label{fig:pipeline}
\end{figure}

No existing world-model dataset provides the structured supervision our method requires: each training sample must contain a video clip, camera intrinsics and extrinsics, a per-frame binary mask identifying the moving subject, and a multi-point trajectory describing its motion.
We build an automatic pipeline that extracts these \emph{(video, camera, mask, trajectory)} tuples from unlabeled video at scale, using only off-the-shelf vision models and physical-plausibility filtering.
Figure~\ref{fig:pipeline} illustrates the end-to-end flow.

\paragraph{Camera estimation.}
For videos lacking camera annotations (e.g., WISA-80K), we run ViPE to recover per-frame intrinsics and $\mathrm{SE}(3)$ poses at metric scale.
For SpatialVID-HQ, camera parameters are provided as normalized intrinsics and world-to-camera quaternion--translation pairs; we convert these to pixel-space intrinsics and world-to-camera $4{\times}4$ matrices, then Slerp-interpolate the sparse annotation frames (sampled at $\lfloor\text{fps}/5\rfloor$ intervals) to the full frame rate.

\paragraph{Subject discovery.}
The central challenge is identifying \emph{which} object to track---videos may contain dozens of moving entities, and only a subset exhibit the kind of coherent, spatially significant motion suitable for trajectory training.
We employ two complementary strategies depending on the data source:

\begin{itemize}[leftmargin=*,nosep]
\item \textbf{VLM-guided discovery} (WISA-80K).
We sample 5 frames uniformly from each clip and query a vision-language model (Qwen2.5-VL-8B) with a structured prompt asking it to identify the most salient moving subject and describe its appearance.
The VLM response is parsed into a text query, which is passed to GroundingDINO to produce a bounding box localized in the video.
This approach requires no category-specific priors and naturally adapts to open-vocabulary scenes.

\item \textbf{Multi-frame tracklet matching} (SpatialVID-HQ).
SpatialVID-HQ provides binary dynamic-region masks (\texttt{dyn\_masks}) that mark \emph{all} moving pixels per frame, but do not distinguish individual objects.
In crowded scenes a single frame may contain 30--300 connected components of varying size; na\"ively selecting the largest component yields multi-person merged blobs (area ratio $>0.3$) rather than individual entities.
We instead perform cross-frame association: on each annotated frame, we extract connected components filtered to $0.1\%$--$30\%$ of image area, then greedily match components across frames by centroid proximity (threshold: $15\%$ of image diagonal).
The resulting \emph{tracklets} capture per-object temporal persistence; we score each tracklet by $\text{score} = n_\text{frames} \times s_\text{area} \times s_\text{coherence}$, where $s_\text{area}$ peaks for objects occupying $0.5\%$--$15\%$ of the frame and $s_\text{coherence}$ penalizes erratic centroid jumps.
The top-scoring tracklet yields the target frame, bounding box, centroid, and a per-pixel component mask for SAM2 initialization.
\end{itemize}

\paragraph{Tracking and trajectory extraction.}
Given the discovered subject (bounding box and, when available, centroid and component mask), we initialize SAM2 video segmentation with a compound prompt: per-pixel mask logits provide the strongest initialization signal, the centroid serves as a positive point to anchor identity, and the bounding box acts as a spatial fallback.
This triple-prompt strategy substantially reduces identity switches in crowded scenes compared to box-only prompting.
SAM2 propagates bidirectionally from the prompt frame, producing per-frame binary masks.

We then identify the optimal 97-frame window by sliding over the SAM2 output and selecting the interval with maximum mask coverage ($\geq 30\%$ of frames must contain a valid subject mask).
Within this window, we verify that the subject exhibits meaningful displacement: centroid net displacement must exceed a minimum threshold, filtering out near-stationary objects whose ``motion'' is merely camera-induced parallax.

Finally, CoTracker3 tracks 20 query points seeded from the subject mask (one centroid plus 19 uniformly sampled interior points), producing dense multi-point trajectories over 97 frames.
The center-point trajectory serves as the primary training signal; the remaining 19 tracks provide auxiliary supervision for spatial extent.

\begin{figure}[t]
  \centering
  \includegraphics[width=\linewidth]{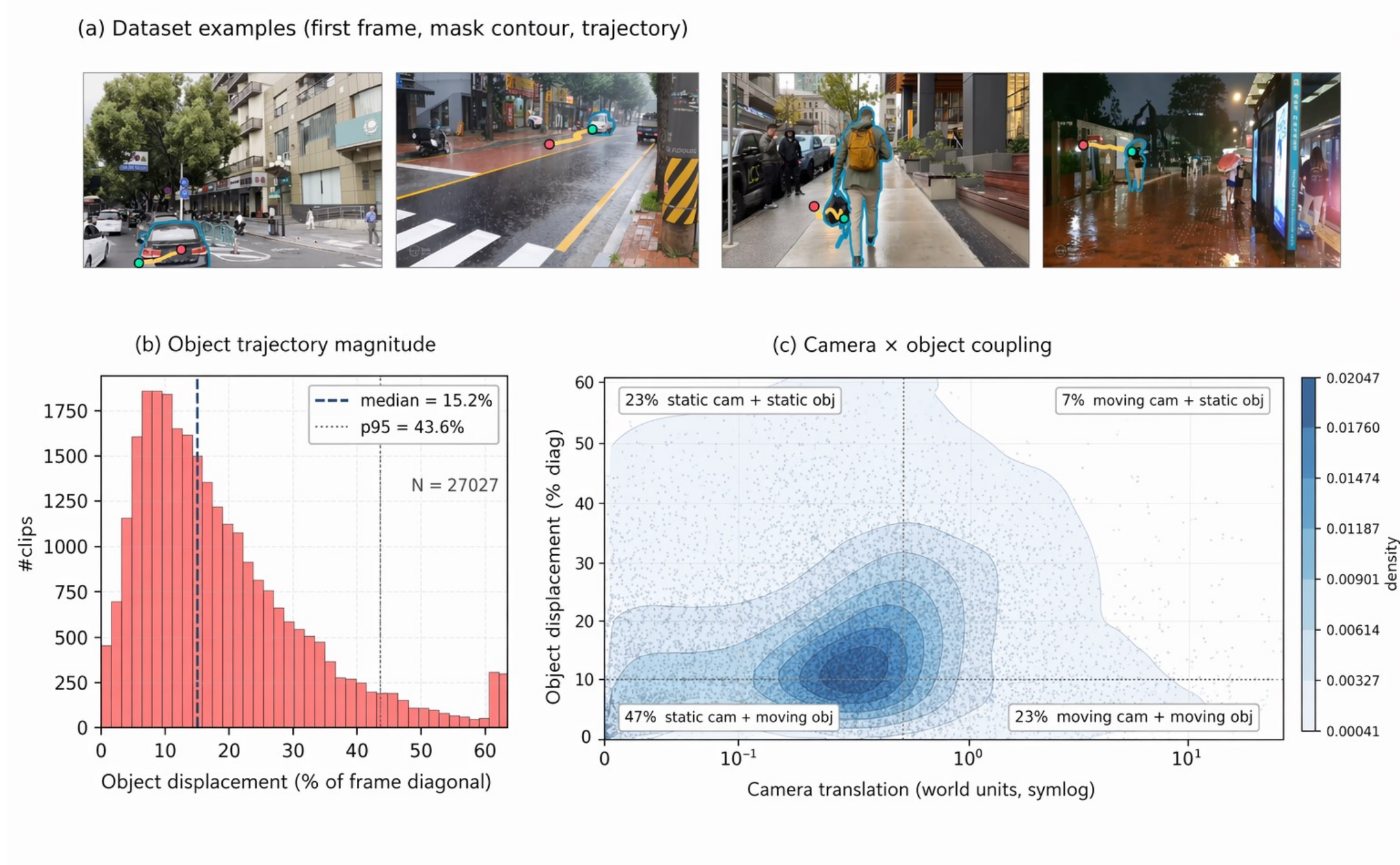}
  \caption{\textbf{Curated training set statistics} ($N{=}27{,}027$ clips after filtering).
  \textbf{(a)} Representative samples with the first frame, the SAM2 mask contour of the selected subject (blue), and the multi-point trajectory overlay (start in green, end in red, path in yellow). Subjects range from vehicles and pedestrians to pushed or carried objects under diverse weather and lighting.
  \textbf{(b)} Distribution of object displacement magnitude, measured as the net 2D displacement of the subject centroid across the 97-frame window, normalized by the frame diagonal. The distribution is right-skewed with median $15.2\%$ and $p_{95}{=}43.6\%$, covering small to large object motions.
  \textbf{(c)} Joint distribution of camera translation (world units, symlog axis) and object displacement (\% diagonal). Using thresholds of $0.5$ world-units for camera and $10\%$ diagonal for object, four regimes partition the dataset: $47\%$ static-cam / moving-obj (purely object-centric), $23\%$ static-cam / static-obj, $7\%$ moving-cam / static-obj (pure ego-motion), and $23\%$ moving-cam / moving-obj, the WorldCraft-specific regime that demands composable control and is absent from most existing trajectory datasets.}
  \label{fig:dataset-pipeline-stats}
\end{figure}

\paragraph{Training data summary.}
Table~\ref{tab:data} summarizes the curated datasets.
All videos are standardized to 30\,fps and 97 frames (${\approx}3.2$\,s).
Camera parameters are converted to a unified format of per-frame $3{\times}3$ intrinsic and $4{\times}4$ world-to-camera matrices.
Video latents are pre-cached through the HunyuanVideo VAE encoder to accelerate training. Figure~\ref{fig:dataset-pipeline-stats} reports sample-level statistics of the $27{,}027$ filtered clips: object-displacement magnitude and the joint distribution of camera and object motion.

\begin{table}[t]
  \caption{Training data statistics. ``Camera source'' indicates whether camera parameters are estimated by our pipeline or provided by the dataset. ``Subject method'' indicates how the target object is identified.}
  \label{tab:data}
  \centering
  \small
  \begin{tabular}{lccccc}
    \toprule
    Dataset & Raw & After curation & Camera source & Subject method & Traj.\ points \\
    \midrule
    WISA-80K      & 80K  & 8.5K  & ViPE (estimated) & VLM + GroundingDINO & 20 \\
    SpatialVID-HQ & 100K & 24K   & Provided (metric) & Tracklet matching   & 20 \\
    OmniWorld     & ---  & 2K    & Provided          & Provided            & --- \\
    \midrule
    \textbf{Total} & ---  & \textbf{${\sim}$34.5K} & --- & --- & --- \\
    \bottomrule
  \end{tabular}
\end{table}

\section{Spatial-Pathway LoRA: motivation, design, and ablations}
\label{sec:splora-ablation}

\paragraph{Shared spatial-control pathway.}
Camera navigation and object trajectories act at different spatial scales but impose the same type of requirement on the video DiT: an external motion signal must change where visual content is rendered. Camera actions produce global scene transformations, whereas object trajectories produce local transformations of selected instances. In WorldPlay, camera actions are processed by the action encoder \texttt{action\_in} and Projective Positional Encoding projections \texttt{prope\_proj}. We therefore test whether trajectory control can reuse this existing spatial-control pathway rather than introduce broad changes to attention routing or channel mixing.

\paragraph{Evidence from full-parameter adaptation.}
We first fine-tune the full backbone for trajectory control and measure the relative change of each parameter group,
\begin{equation}
\Delta_\text{rel}^{(l)} =
\frac{\lVert \mathbf{W}^{(l)}_\text{ft} - \mathbf{W}^{(l)}_\text{base} \rVert_F}
{\lVert \mathbf{W}^{(l)}_\text{base} \rVert_F}.
\label{eq:appendix-weight-change}
\end{equation}
The optimizer concentrates its largest changes on \texttt{action\_in} and \texttt{prope\_proj}: these modules change by $10$-$25{\times}$ more than attention Q/K/V and MLP layers. Table~\ref{tab:delta} reports the highest-change parameters and shows that no attention Q/K/V, attention-output projection, or MLP parameter appears among the top 30. This pattern indicates that adapting the mapping from spatial conditioning to feature positions is more important than relearning general semantic routing.

\paragraph{Spatial-Pathway LoRA.}
Based on this observation, we restrict LoRA~\citep{lora} updates to the spatial-control pathway:
\begin{equation}
\mathbf{W}'^{(l)} =
\mathbf{W}^{(l)} + \mathbf{B}^{(l)}\mathbf{A}^{(l)}\frac{\alpha}{r},
\qquad
l \in \{\texttt{action\_in},\;\texttt{prope\_proj}^{(1..L)}\},
\label{eq:appendix-splora}
\end{equation}
while freezing attention, MLP, and all remaining backbone parameters. This produces a ${\sim}50$M-parameter adapter, approximately $0.6\%$ of the 8B backbone. The frozen base weights retain the pretrained camera response, while the low-rank update learns how trajectory conditions perturb the same spatial pathway.

\paragraph{Layer selection and training curriculum.}
Table~\ref{tab:layer-ablation} evaluates the two design decisions that follow from this formulation. In Table~\ref{tab:layer-ablation}(a), full fine-tuning obtains the lowest trajectory error but substantially degrades camera fidelity. Conventional LoRA on Q/K/V and MLP layers, and wider variants that additionally adapt V and MLP layers, use more trainable parameters without improving the TE-RPE trade-off. Spatial-Pathway LoRA provides the best balance between trajectory accuracy and inherited camera control. Table~\ref{tab:layer-ablation}(b) then holds the adapter fixed and compares training schedules. Static-BI $\to$ Dynamic-AR best preserves both capabilities: static bidirectional training first establishes trajectory following, and dynamic autoregressive training transfers it to camera-controlled long-horizon generation.

\begin{table}[ht]
  \caption{\textbf{Adaptation and training ablation.}
  (a)~Spatial-Pathway LoRA, provides the best TE-RPE trade-off with the fewest trainable parameters.
  (b)~The Static-BI $\to$ Dynamic-AR curriculum gives the best joint preservation of trajectory control and camera fidelity.}
  \centering
  \small
  \setlength{\tabcolsep}{6pt}
  \begin{tabular}{l | c cc}
    \toprule
    \textbf{Configuration} & \textbf{\#Params} & \textbf{TE}\,\down & \textbf{RPE}\textsubscript{rot}\,\down \\
    \midrule
    \multicolumn{4}{l}{\textit{(a) Layer selection, trained with Static-BI $\to$ Dynamic-AR}} \\[2pt]
    \textbf{Spatial-Pathway LoRA (ProPE + action)} & ${\sim}$50M  & 38.90 & \textbf{0.131} \\
    Spatial pathway + V + MLP (blocks 28-42)      & ${\sim}$120M & 46.60 & 0.136 \\
    Spatial pathway + V + MLP (all blocks)         & ${\sim}$180M & 47.65 & 0.169 \\
        Q/K/V + MLP (conventional LoRA)                & ${\sim}$200M & 49.43 & 0.139 \\
    Full fine-tune                                 & 8B           & \textbf{37.20} & 0.237 \\
    \midrule
    \multicolumn{4}{l}{\textit{(b) Training strategy, using Spatial-Pathway LoRA}} \\[2pt]
    Dynamic data, AR attention                     & ${\sim}$50M & 46.63          & 0.170 \\
    Dynamic BI $\to$ Dynamic AR                  & ${\sim}$50M & \textbf{38.25} & 0.177 \\
    \textbf{Static BI $\to$ Dynamic AR}            & ${\sim}$50M & 38.90          & \textbf{0.131} \\
    \bottomrule
  \end{tabular}
  \vspace{-3mm}
  \label{tab:layer-ablation}
\end{table}

\subsection{Camera-trajectory interaction analysis}

Beyond the weight-space and output-level results above, we analyze how trajectory adaptation interacts with the inherited camera representation. Rather than running full video generation, each activation experiment performs a single denoising forward pass and probes the \texttt{prope\_proj} output across all 54 DiT double-stream blocks. Comparisons use the same denoising timestep and noise realization while varying only the camera and trajectory conditions. This controlled setup isolates their relative activation effects, and the reported trends remain stable across the tested timestep choices.

\begin{figure}[t]
  \centering
  \includegraphics[width=\linewidth]{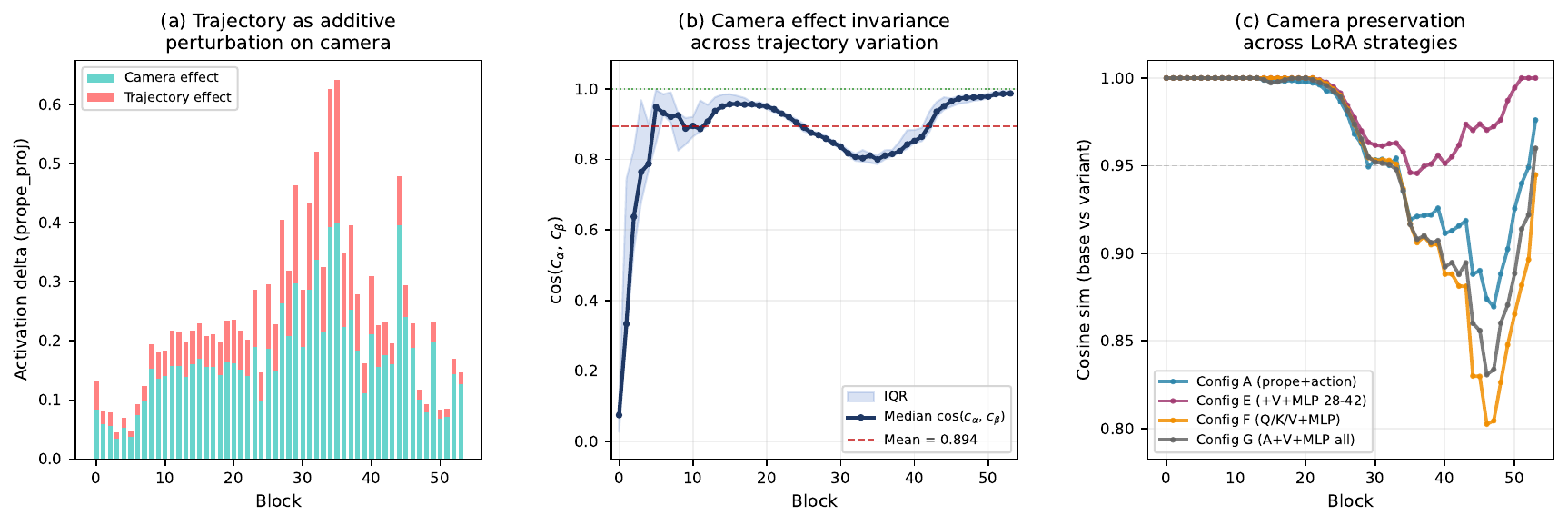}
  \caption{Activation-level analysis of camera\,-\,trajectory interaction.
  \textbf{(a)}~Trajectory effect is a small additive perturbation on top of the camera signal across all blocks (stacked bars on \texttt{prope\_proj}).
  \textbf{(b)}~Counterfactual camera invariance: $\cos(\mathbf{c}_\alpha, \mathbf{c}_\beta)$ per block, where $\mathbf{c}$ is the camera effect vector measured under two different trajectories. Mean = 0.89, confirming that trajectory variation does not alter the camera effect direction.
  \textbf{(c)}~Layer selection ablation on camera preservation: per-block cosine similarity averaged across all hooked layer types (\texttt{img\_mod}, \texttt{prope\_proj}, Q/K/V, MLP, \texttt{img\_attn\_proj}) between base model and each LoRA variant under camera-only input. Pathway-selective adaptation (prope+action) preserves camera activations far better than Q/K/V LoRA, which causes significant degradation in mid-to-late blocks.}
  \label{fig:analysis}
\end{figure}

\paragraph{How large is the trajectory signal?}
We perform four forward passes with different input conditions (baseline, camera-only, trajectory-only, combined) and decompose the per-layer activation delta.
Figure~\ref{fig:analysis}(a) shows the trajectory signal as a thin additive layer on top of the camera signal across all 54 blocks.
The stacked-bar structure shows that LoRA preserves the dominant camera representation and adds a smaller trajectory-specific perturbation (mean trajectory-to-camera energy ratio $=0.42$).

\paragraph{Is the camera effect direction preserved?}
To directly test whether camera control is preserved under trajectory conditioning, we employ a $2 \times 2$ counterfactual design: two camera poses (A, B) crossed with two trajectories ($\alpha$, $\beta$).
For each layer, the camera effect vector is $\mathbf{c}_\alpha = h(B, \alpha) - h(A, \alpha)$. Camera invariance is measured by $\cos(\mathbf{c}_\alpha, \mathbf{c}_\beta)$.
A value near 1 indicates that the direction of the camera effect is unchanged by trajectory variation.
Critically, this probe is evaluated on the \texttt{prope\_proj} output, which encodes continuous camera pose (viewmat) and is the layer that shares parameters with the trajectory adapter.
Corruption of the continuous camera pathway would rotate the camera-effect direction. A small but misaligned perturbation would preserve its magnitude while reducing directional agreement.
Figure~\ref{fig:analysis}(b) shows that from block 5 onward, $\cos(\mathbf{c}_\alpha, \mathbf{c}_\beta)$ is consistently above 0.85, with a cross-block mean of 0.89.
This demonstrates that the camera effect direction is highly stable regardless of trajectory input, confirming that pathway-selective LoRA achieves an asymmetric decoupling: camera control is preserved while trajectory control is added.

\noindent\textbf{Do camera and trajectory share a feature subspace?}

\begin{figure}[t]
  \centering
  \includegraphics[width=\linewidth]{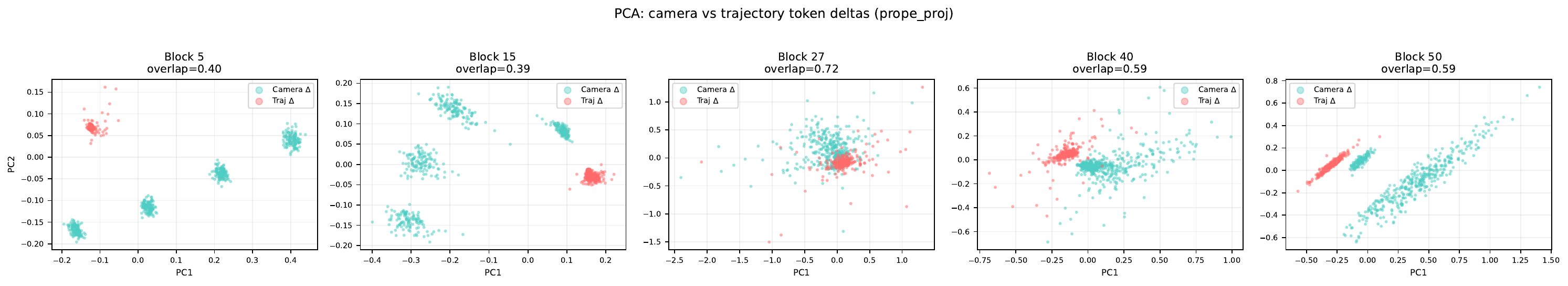}

\caption{\textbf{Shared control subspace.} 2D PCA projection of token-level camera-control updates $\mathbf{u}$ (blue) and trajectory-control updates $\mathbf{v}$ (red) at the peak block. The two distributions are aligned along the same principal directions rather than forming orthogonal subspaces, indicating that trajectory control is injected within the camera-compatible spatial-control subspace.}
  \label{fig:pca-subspace}
  \vspace{-3mm}
\end{figure}

A deeper question is \emph{why} trajectory control can be added without destructively interfering with the original camera-control ability.
We analyze token-level activation updates induced by camera control, $\mathbf{u}=\mathbf{h}_\text{cam}-\mathbf{h}_\text{base}$, and trajectory control, $\mathbf{v}=\mathbf{h}_\text{cam+traj}-\mathbf{h}_\text{cam}$.
Using PCA and cosine-based subspace overlap from principal angles, Figure~\ref{fig:pca-subspace} shows that the two updates are most aligned in the middle layers, where spatial layout and geometric control are primarily represented.

Early and late layers show lower overlap, indicating that the two signals remain sufficiently distinguishable for low-level visual encoding and final rendering.
Together with the preserved camera accuracy in Table~\ref{tab:layer-ablation}, this supports our claim that WorldCraft adds object-level control by reusing the existing spatial pathway while avoiding destructive interference with camera control.

\begin{table}[h]
  \centering
  \small
  \caption{Top-30 parameters ranked by relative weight change $\Delta_\text{rel} = \lVert \mathbf{W}_\text{ft} - \mathbf{W}_\text{base} \rVert_F / \lVert \mathbf{W}_\text{base} \rVert_F$ after full-parameter trajectory fine-tuning of WorldPlay (8B).
  The ranking is dominated by \texttt{action\_in} (ranks 1-2), \texttt{prope\_proj} (22 of the top 30 rows), and the \texttt{final\_layer} adaLN modulation, confirming that the optimizer concentrates updates on the spatial-control pathway.
  No attention Q/K/V, attention-output projection, or MLP parameter appears in the top 30.}
  \label{tab:delta}
  \begin{tabular}{rcccrll}
    \toprule
    Rank & $\Delta_\text{rel}$ & $\lVert\Delta\mathbf{W}\rVert_F$ & $\lVert\mathbf{W}_\text{base}\rVert_F$ & Params & Category & Parameter \\
    \midrule
    1 & 0.9619 & 0.2835 & 0.2947 & 4{,}194{,}304 & action & \texttt{action\_in.mlp.W} \\
    2 & 0.1725 & 0.0037 & 0.0213 & 2{,}048 & action & \texttt{action\_in.mlp.b} \\
    3 & 0.1371 & 0.1214 & 0.8860 & 4{,}194{,}304 & prope & \texttt{blk53.prope.W} \\
    4 & 0.0996 & 1.7929 & 18.0008 & 8{,}388{,}608 & other & \texttt{final.adaLN.W} \\
    5 & 0.0991 & 0.0017 & 0.0171 & 2{,}048 & prope & \texttt{blk53.prope.b} \\
    6 & 0.0945 & 0.1342 & 1.4196 & 4{,}194{,}304 & prope & \texttt{blk52.prope.W} \\
    7 & 0.0903 & 0.0018 & 0.0203 & 2{,}048 & prope & \texttt{blk52.prope.b} \\
    8 & 0.0883 & 0.0018 & 0.0206 & 2{,}048 & prope & \texttt{blk51.prope.b} \\
    9 & 0.0827 & 0.0018 & 0.0214 & 2{,}048 & prope & \texttt{blk50.prope.b} \\
    10 & 0.0824 & 0.1103 & 1.3388 & 4{,}194{,}304 & prope & \texttt{blk3.prope.W} \\
    11 & 0.0820 & 0.0981 & 1.1956 & 4{,}194{,}304 & prope & \texttt{blk2.prope.W} \\
    12 & 0.0768 & 0.1212 & 1.5794 & 4{,}194{,}304 & prope & \texttt{blk51.prope.W} \\
    13 & 0.0765 & 0.0018 & 0.0240 & 2{,}048 & prope & \texttt{blk32.prope.b} \\
    14 & 0.0759 & 0.1171 & 1.5435 & 4{,}194{,}304 & prope & \texttt{blk5.prope.W} \\
    15 & 0.0758 & 0.0018 & 0.0240 & 2{,}048 & prope & \texttt{blk30.prope.b} \\
    16 & 0.0755 & 0.1239 & 1.6413 & 4{,}194{,}304 & prope & \texttt{blk6.prope.W} \\
    17 & 0.0753 & 0.0019 & 0.0247 & 2{,}048 & prope & \texttt{blk29.prope.b} \\
    18 & 0.0753 & 0.0018 & 0.0240 & 2{,}048 & prope & \texttt{blk31.prope.b} \\
    19 & 0.0747 & 0.0020 & 0.0268 & 2{,}048 & prope & \texttt{blk26.prope.b} \\
    20 & 0.0745 & 0.1182 & 1.5865 & 4{,}194{,}304 & prope & \texttt{blk4.prope.W} \\
    21 & 0.0739 & 0.0019 & 0.0251 & 2{,}048 & prope & \texttt{blk28.prope.b} \\
    22 & 0.0735 & 0.1269 & 1.7267 & 4{,}194{,}304 & prope & \texttt{blk8.prope.W} \\
    23 & 0.0730 & 0.0020 & 0.0273 & 2{,}048 & prope & \texttt{blk25.prope.b} \\
    24 & 0.0730 & 0.1247 & 1.7075 & 4{,}194{,}304 & prope & \texttt{blk7.prope.W} \\
    25 & 0.0725 & 0.0021 & 0.0286 & 2{,}048 & prope & \texttt{blk24.prope.b} \\
    26 & 0.0723 & 0.0019 & 0.0266 & 2{,}048 & prope & \texttt{blk27.prope.b} \\
    27 & 0.0723 & 0.1304 & 1.8030 & 4{,}194{,}304 & prope & \texttt{blk9.prope.W} \\
    28 & 0.0713 & 0.0021 & 0.0298 & 2{,}048 & prope & \texttt{blk23.prope.b} \\
    29 & 0.0697 & 0.0017 & 0.0247 & 2{,}048 & prope & \texttt{blk34.prope.b} \\
    30 & 0.0697 & 0.1138 & 1.6325 & 4{,}194{,}304 & prope & \texttt{blk50.prope.W} \\
    \bottomrule
  \end{tabular}
\end{table}

\section{Supplementary evaluations}
\label{sec:supp-joint}

\paragraph{Protocol.}
\rev{The following tables report each baseline on the control outputs defined by its interface. ``\na'' denotes an unavailable output. Trajectory methods are evaluated for object following, camera-controlled world models for navigation and visual reconstruction, and WorldCraft for both within the same rollout.}

\subsection{Composable camera-object control}
\label{sec:composable-supp}

The Composable set contains 45 held-out clips, evenly divided among small, medium, and large camera rotations. Each example specifies both a camera trajectory and a selected-object trajectory. Table~\ref{tab:composable-61} reports the 61-frame setting. The specialist baselines remain competitive on the axis they support, but only WorldCraft executes both requested actions in one rollout.

\begin{table}[t]
  \centering
  \scriptsize
  \caption{Composable camera-object control at 61 frames, averaged over 45 held-out clips. ``\na'' denotes that the method has no interface for the corresponding output. Joint success requires both control axes in the same rollout.}
  \label{tab:composable-61}
  \resizebox{\linewidth}{!}{
  \begin{tabular}{llccccccccc}
    \toprule
    Type & Method & TE\down & DINO\up & PSNR\up & SSIM\up & LPIPS\down & RPE$_\text{rot}$\down & RPE$_\text{trans}$\down & RPE$_\text{cam}$\down & Joint \\
    \midrule
    Traj. & DragAnything & 39.76 & 0.7482 & \na & \na & \na & \na & \na & \na & \na \\
    Traj. & Wan-Move & 35.42 & 0.7516 & \na & \na & \na & \na & \na & \na & \na \\
    Cam. & GameCraft & \na & \na & 13.15 & 0.3180 & 0.5158 & 0.173 & 0.0155 & 0.0165 & \na \\
    Cam. & WorldPlay & \na & \na & 14.37 & 0.3651 & 0.4370 & 0.133 & 0.0159 & 0.0179 & \na \\
    Joint & \textbf{WorldCraft} & \textbf{32.46} & \textbf{0.8799} & \textbf{14.46} & \textbf{0.3683} & \textbf{0.4306} & \textbf{0.114} & \textbf{0.0146} & \textbf{0.0163} & \yes \\
    \bottomrule
  \end{tabular}
  }
\end{table}

Table~\ref{tab:composable-253} extends the same joint protocol to 253 frames ($\approx10.5$ seconds). Fixed-window trajectory baselines do not define a streaming camera-plus-object rollout at this horizon, whereas WorldCraft continues to expose both action channels. Relative to the camera-only WorldPlay baseline, WorldCraft retains competitive or improved camera metrics while additionally executing the object trajectory.

\begin{table}[t]
  \centering
  \scriptsize
  \caption{Long-horizon composable control at 253 frames. Fixed-window trajectory baselines do not define a streaming joint rollout and are therefore omitted.}
  \label{tab:composable-253}
  \resizebox{\linewidth}{!}{
  \begin{tabular}{llccccccccc}
    \toprule
    Type & Method & TE\down & DINO\up & PSNR\up & SSIM\up & LPIPS\down & RPE$_\text{rot}$\down & RPE$_\text{trans}$\down & RPE$_\text{cam}$\down & Joint \\
    \midrule
    Cam. & GameCraft & \na & \na & 12.03 & 0.2964 & 0.6518 & 0.196 & 0.0213 & 0.0226 & \na \\
    Cam. & WorldPlay & \na & \na & 12.94 & 0.3385 & 0.5521 & 0.142 & 0.0229 & 0.0210 & \na \\
    Joint & \textbf{WorldCraft} & \textbf{43.25} & \textbf{0.8028} & \textbf{13.23} & \textbf{0.3458} & \textbf{0.5378} & \textbf{0.134} & \textbf{0.0194} & \textbf{0.0186} & \yes \\
    \bottomrule
  \end{tabular}
  }
\end{table}

\subsection{Held-out DROID robot manipulation}
\label{sec:droid}

\paragraph{Protocol.}
We additionally evaluate transfer to robot manipulation using the DROID subset of OmniWorld~\citep{droid,omniworld}. DROID videos are captured by a wrist-mounted camera on the robot end-effector, so camera and manipulated-object motion are coupled by construction. We annotate 2D trajectories for the robot arm and manipulated object with CoTracker, yielding paired object and camera ground truth for 50 videos of 125 frames each. The DROID evaluation videos are excluded from WorldCraft training, and we perform no DROID-specific adaptation.

\begin{table}[b]
  \centering
  \scriptsize
  \caption{Held-out DROID robot-manipulation control (50 videos, 125 frames). Each method is evaluated on the outputs defined by its control interface. ``\na'' denotes an unavailable output.}
  \label{tab:droid}
  \resizebox{\linewidth}{!}{
  \begin{tabular}{llcccccccc}
    \toprule
    Type & Method & TE\down & DINO\up & RPE$_\text{rot}$\down & RPE$_\text{trans}$\down & RPE$_\text{cam}$\down & PSNR\up & SSIM\up & LPIPS\down \\
    \midrule
    Traj. & Wan-Move & 45.39 & 0.7011 & \na & \na & \na & \na & \na & \na \\
    Cam. & WorldPlay & \na & \na & 0.145 & 0.0225 & 0.0203 & 12.95 & 0.4213 & 0.4703 \\
    Joint & \textbf{WorldCraft} & \textbf{37.77} & \textbf{0.7970} & \textbf{0.127} & \textbf{0.0194} & \textbf{0.0175} & \textbf{13.38} & \textbf{0.4433} & \textbf{0.4334} \\
    \bottomrule
  \end{tabular}
  }
\end{table}

\paragraph{Results.}
\rev{WorldCraft transfers to a held-out manipulation domain while jointly controlling the wrist-camera motion and selected entities. It lowers trajectory error relative to Wan-Move (37.77 versus 45.39) and improves identity consistency (0.797 versus 0.701). Despite executing the additional object action, it also improves all reported camera-pose and visual-quality metrics over WorldPlay. The experiment demonstrates transfer to unseen manipulation scenes with coupled camera-object motion. It evaluates kinematic control and does not measure causal contact dynamics.}

\end{document}